\documentclass[conference,a4paper]{IEEEtran}

\usepackage[final]{graphicx}
\usepackage[reqno]{amsmath}
\usepackage{amssymb}
\usepackage{subfig}
\usepackage{epstopdf}
\usepackage{xcolor}
\usepackage{booktabs}

\begin{document}

\sloppy

\title{Deep Neural Network Architectures for Modulation Classification
}
\author{\IEEEauthorblockN{Xiaoyu Liu, Diyu Yang, and Aly El Gamal}
 \IEEEauthorblockA{School of Electrical and Computer Engineering \\Purdue University\\ Email: \{liu1962, yang1467, elgamala\}@purdue.edu}}

\maketitle

\begin{abstract}
In this work, we investigate the value of employing deep learning for the task of wireless signal modulation recognition. Recently in~\cite{conv}, a framework has been introduced by generating a dataset using GNU radio that mimics the imperfections in a real wireless channel, and uses 10 different modulation types. Further, a convolutional neural network (CNN) architecture was developed and shown to deliver performance that exceeds that of expert-based approaches. Here, we follow the framework of~\cite{conv} and find deep neural network architectures that deliver higher accuracy than the state of the art. We tested the architecture of~\cite{conv} and found it to achieve an accuracy of approximately 75\% of correctly recognizing the modulation type. We first tune the CNN architecture of~\cite{conv} and find a design with four convolutional layers and two dense layers that gives an accuracy of approximately 83.8\% at high SNR. We then develop architectures based on the recently introduced ideas of Residual Networks (ResNet~\cite{resnet}) and Densely Connected Networks (DenseNet~\cite{densenet}) to achieve high SNR accuracies of approximately 83.5\% and 86.6\%, respectively. Finally,  we introduce a Convolutional Long Short-term Deep Neural Network (CLDNN~\cite{CLDNN}) to achieve an accuracy of approximately 88.5\% at high SNR. 

\end{abstract}

\section{Introduction}

Signal modulation is an essential process in wireless communication systems. Modulation recognition tasks are generally used for both signal detection and demodulation. The signal transmission can be smoothly processed only when the signal receiver demodulates the signal correctly. However, with the fast development of wireless communication techniques and more high-end requirements, the number of modulation methods and parameters used in wireless communication systems is increasing rapidly. The problem of how to recognize modulation methods accurately is hence becoming more challenging. 

Traditional modulation recognition methods usually require prior knowledge of signal and channel parameters, which can be inaccurate under mild circumstances and need to be delivered through a separate control channel. Hence, the need for autonomous modulation recognition arises in wireless systems, where modulation schemes are expected to change frequently as the environment changes. This leads to considering new modulation recognition methods using deep neural networks.

Deep Neural Networks (DNN) have played a significant role in the research domain of video, speech and image processing in the past few years. Recently the idea of deep learning has been introduced to the area of communications by applying convolutional neural networks (CNN) to the task of radio modulation recognition \cite{conv}. 

The Convolutional Neural Network (CNN) has been recently identified as a powerful tool in image classification and voice signal processing. There have also been successful attempts to apply this method in other areas such as natural language processing and video detection. Based on its supreme performance in feature extraction, a simple architecture of CNN was introduced in~\cite{conv} for distinguishing between 10 different modulations. Simulation results show that CNN not only demonstrates better accuracy results, but also provides more flexibility compared to current day expert-based approaches \cite{conv}. 
\begin{figure}[htb]
\centering
\includegraphics[width=0.8\columnwidth]{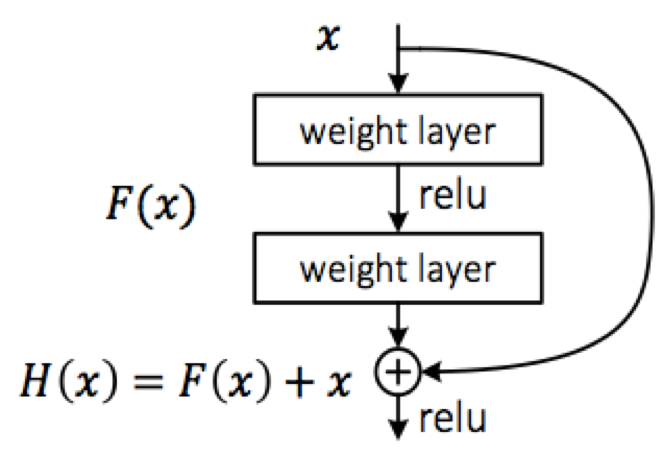}
\caption{A building block of ResNet.}
\label{fig:res}
\end{figure} 
However, CNN has been challenged with problems like vanishing or exploding gradients, and accuracy degradation after reaching a certain network depth. Attempts have been made to address the above issues. Most notably, Residual Networks (ResNet) \cite{resnet} and Densely Connected Networks (DenseNet) \cite{densenet} were recently introduced to strengthen feature propagation in the neural network by creating shortcut paths between different layers in the network. A building block of a residual learning network can be expressed using the equation in Figure~\ref{fig:res}, where $x$ and $H$($x$) are input and output of the block respectively, and $F$ is the residual mapping function to be trained. Since it may be hard to learn the mapping $H$($x$) = $x$, this block learns the residual mapping $F$($x$) = $H$($x$) - $x$, which can be easier to learn~\cite{resnet}. By adding the bypass connection, an identity mapping is created, allowing the deep network to learn simple functions that would have required a shallower network to learn.

Recently, a Convolutional Long Short-term Deep Neural Network (CLDNN) has been introduced in~\cite{CLDNN}, where it combines the architectures of CNN and Long Short-Term Memory (LSTM) into a deep neural network by taking advantage of the complementarity of CNNs, LSTMs, and DNNs \cite{CLDNN}. The LSTM unit is a memory unit of a Recurrent Neural Network (RNN). RNNs are neural networks with memory that are suitable for learning sequence tasks such as speech recognition and handwritten recognition. LSTM optimizes the gradient vanishing problem in RNNs by using a forget gate in its memory cell, which enables the learning of long-term dependencies.

Due to the fact that traditional channel models of the wireless channel may not be accurate, in our experiments, we use the RadioML2016.10b dataset generated in~\cite{conv} as the input dataset. The data is generated in a way that captures various channel imperfections that are present in a real system using GNU radio.
In this paper, we develop architectures of ResNet, DenseNet, and CLDNN for the modulation recognition task.  Using the same dataset generated in~\cite{conv}, we achieve a roughly 13.5\% accuracy improvement at high SNR against the state of the art architecture presented in \cite{conv}. The improvements of accuracy are believed to be achieved by better spatial and temporal feature extraction.

\section{Simulation Setup}\label{sec:systemmodel}
We use the RadioML2016.10b dataset generated in~\cite{conv} as the input data of our research. Details about the generation of this dataset can be found in \cite{datagen}. This dataset contains 10 types of modulations: eight digital and two analog modulations. These consist of BPSK, QPSK, 8PSK, QAM16, QAM64, BFSK, CPFSK, and PAM4 for digital modulations, and WB-FM, and AM-DSB for analog modulations. For digital modulations, the entire Gutenberg works of Shakespeare in ASCII is used, with whitening randomizers applied to ensure equiprobable symbols and bits. For analog modulations, a continuous voice signal is used as input data, which consists primarily of acoustic voice speech with some interludes and off times. The entire dataset is a 128-sample complex time-domain vector generated in GNU radio. 160,000 samples are segmented into training and testing datasets through 128-samples rectangular windowing processing, which is similar to the  windowed continuous acoustic voice signal in voice recognition tasks. The training examples - each consisting of 128 samples - are fed into the neural network in 2*128 vectors with real and imaginary parts separated in complex time samples. The labels in input data include SNR ground truth and the modulation type. The SNR of samples is uniformly distributed from -20dB to +18dB.
All training and testing are done in Keras using Nvidia M60 GPU. We use Adam \cite{adam} from the deep learning library as optimizer in Keras and use Theano as back end. 

\subsection{Evaluation Network}
\begin{figure}[htb]
\centering
\includegraphics[width=0.8\columnwidth]{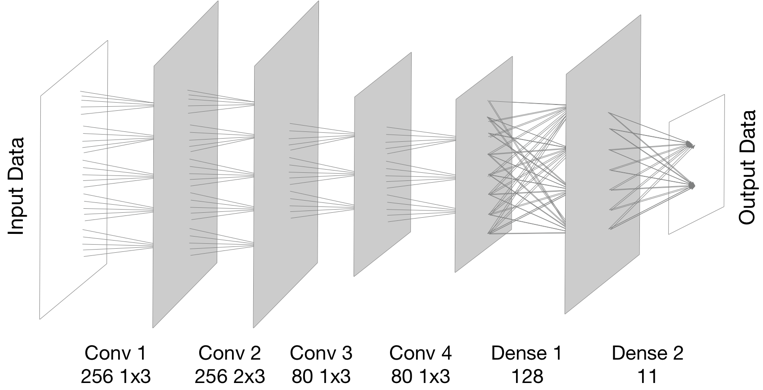}
\caption{Architecture of seven-layer CNN}
\label{fig:1}
\end{figure}

We start with a convolutional neural network architecture similar to the CNN2 network from~\cite{conv}, which performs blind temporal learning using a two-convolutional-layer deep neural network, and achieve accuracy of 75\% at high SNR; a better performance against current day approaches \cite{conv}. Our training is based on several neural network architectures: Convolutional Neural Network (CNN), Densely Connected Convolutional Network (DenseNet) \cite{densenet}, Residual Network (ResNet) \cite{resnet}, and Convolutional Long Short-Term Deep Neural Network (CLDNN)~\cite{CLDNN}. For CNN, we optimized the following hyper-parameters: learning rate, dropout rate, filter size, number of filters per layer and the network depth. We tried different combinations of convolutional layer sequences and filter numbers in each layer to get the best accuracy result. We also develop deeper networks by adding more convolutional layers on the CNN2 model and get the optimal accuracy from the architecture shown in Figure~\ref{fig:1}, where four convolutional layers are followed by two dense layers. The first parameter below each convolutional layer represents the number of filters in that layer, while the second and third numbers show the size of each filter. For the two dense layers, there are 128 and 11 neurons, in order of their depth in the network.

\begin{figure}[htb]
\centering
\includegraphics[width=0.8\columnwidth]{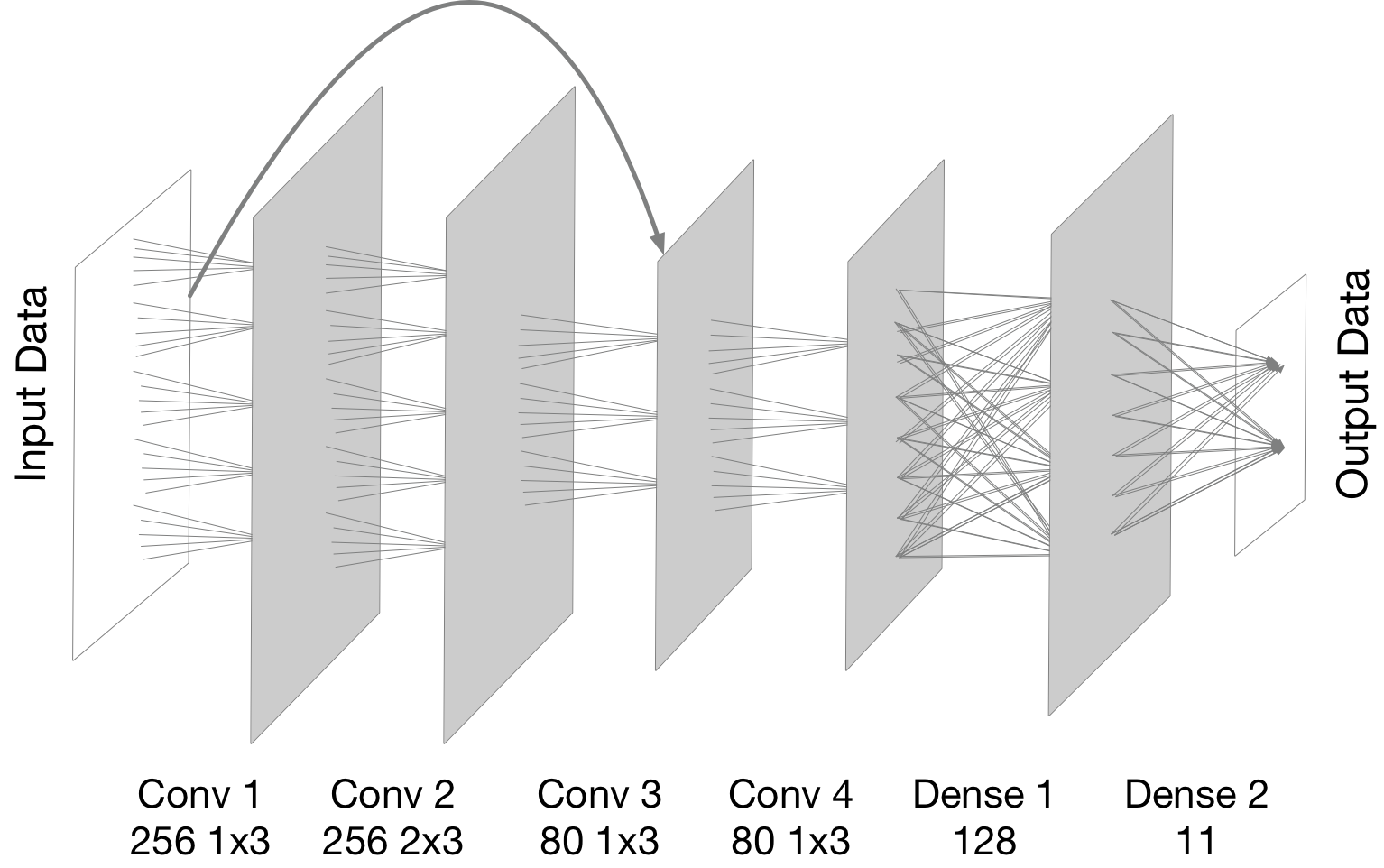}
\caption{Architecture of seven-layer ResNet.}
\label{fig:2}
\end{figure}

Inspired by the winner architecture of ImageNet 2015 \cite{resnet}, we apply the ResNet architecture and test architectures with increasing number of convolutional layers up to 8. We obtain the best classification accuracy from the four-convolutional-layer ResNet architecture shown in Figure~\ref{fig:2}. The output from the first layer is forwarded to the layer two levels deeper. This structure alleviates the gradient vanishing problem by explicitly letting each few stacked layers fit into a residual mapping~\cite{resnet}.  The hyper-parameters are chosen according to the basic observation that we make from simple CNN architectures that having larger filters close to the input layer followed by smaller filters close to the output layer leads to significant accuracy improvement.

\begin{figure}[htb]
\centering
\includegraphics[width=0.8\columnwidth]{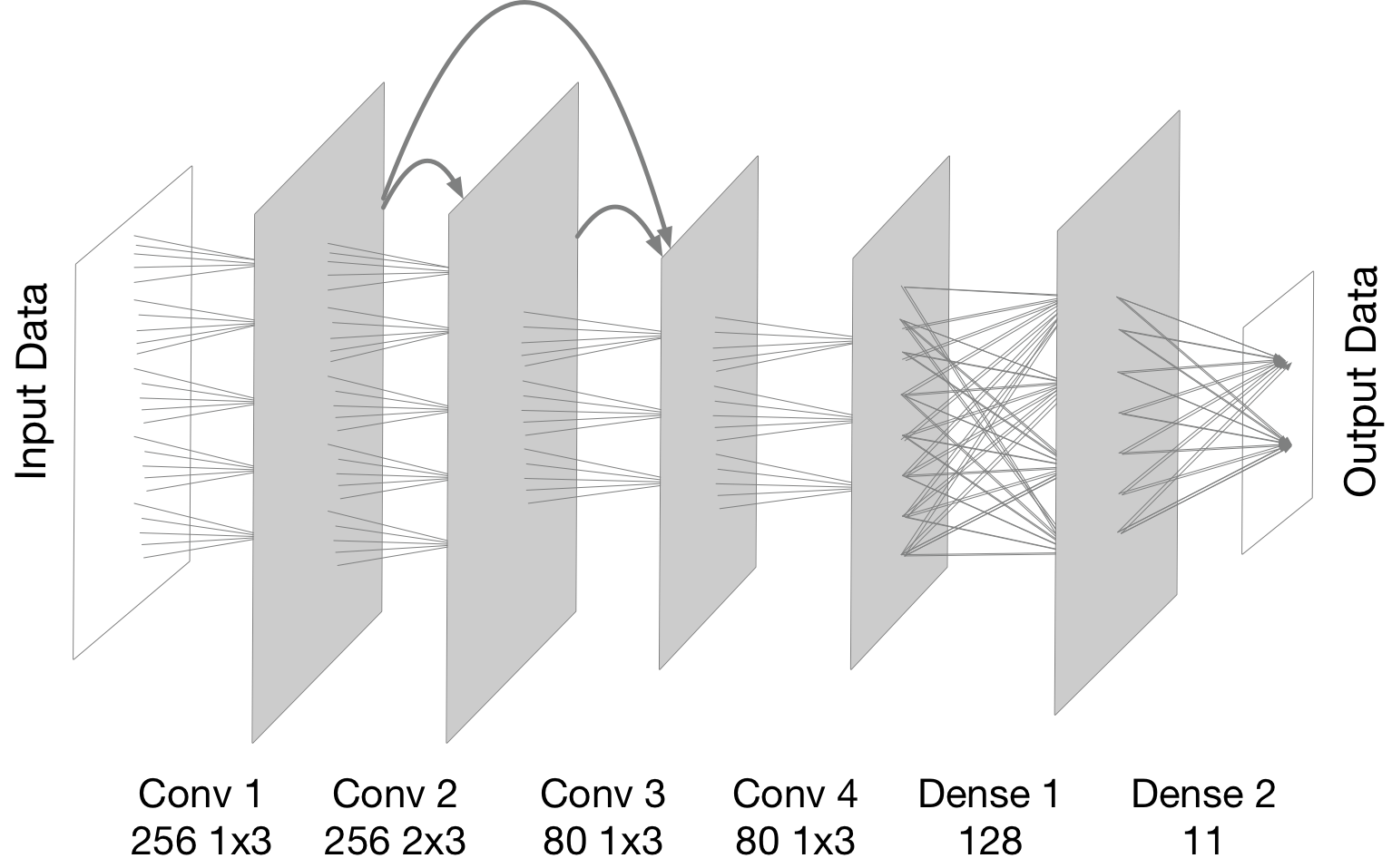}
\caption{Architecture of seven-layer DenseNet.}
\label{fig:3}
\end{figure}

DenseNet further improves the information flow between layers than ResNet does, as each layer obtains additional inputs from all preceding layers and passes on its own feature-maps to all subsequent layers~\cite{densenet}. Our DenseNet architecture is illustrated in Figure~\ref{fig:3}, with four convolutional layers densely connected with each other and the output fed into two dense layers. We set the  parameters of the convolutional layers to achieve the best accuracy. 

\begin{figure}[htb]
\centering
\includegraphics[width=0.8\columnwidth]{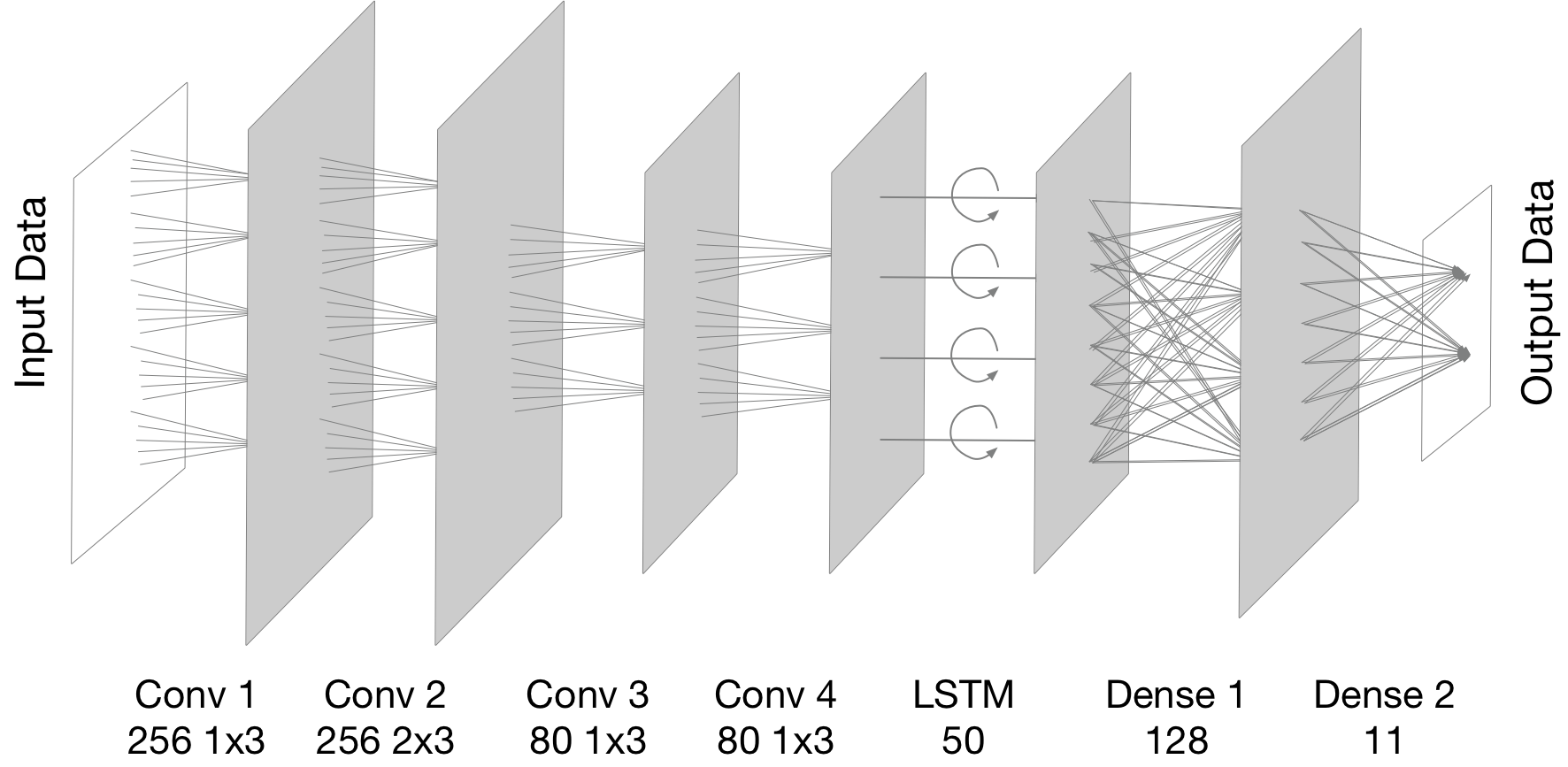}
\caption{Architecture of eight-layer CLDNN.}
\label{fig:lstm}
\end{figure}

We finally propose a CLDNN architecture that includes long short-term memory units. CLDNNs are mainly used in voice processing tasks that involve raw time-domain waveforms~\cite{CLDNN}. It is a combination of CNNs, long short-term memory (LSTM), and deep neural networks (DNN). In our setting, we choose four convolutional layers in CNN, followed by one LSTM layer with 50 computing units and two fully connected DNN layers(see Figure~\ref{fig:lstm}). We tested different CLDNN architectures with different number of memory cells in the LSTM layer. Our experiments show that an LSTM layer with 50 cells gives out the best accuracy result compared to other layer settings.

\subsection{Training Complexity}
The computation time using one NVIDIA M60 GPU for 96000 training examples and 64000 validation and testing examples varies signifiantly for different models. The simplest model with only two convolutional layers in CNN takes approximately 15 seconds per epoch while the CNN with four convolutional layers takes approximately 400 seconds per epoch. We note that a high dropout rate may slow down the training speed but reduces overfitting. In our setting, we set the dropout rate to 0.6, which is higher than the rate used in~\cite{conv}, and the activation function in each hidden layer is a Rectified Linear Unit (ReLU) function. We set patience, the period during which a non-converging validation loss is tolerated, to 10 when there are three and four convolutional layers and get a total training time of around half an hour. When the network becomes deeper, it starts to take more than 10 training epochs for the validation loss to decrease, so setting patience to 20 produces smaller validation loss, which means higher accuracy. To get better results, we set patience to 20 in the remaining models. It takes approximately 1000 seconds per epoch in all three models. The total training time is approximately 70 hours for the DenseNet model, 20 hours for the ResNet model, and 50 hours for the CLDNN model.

\section{Results}

\subsection{Convolutional Neural Network}
We start with a basic two-convolutional-layer neural network, in which two convolutional layers with 256 1x3 filters and 80 2x3 filters, respectively, are followed by two dense layers. We then explore the effect of different filter settings by exchanging filter settings between the two convolutional layers. The performances of networks with different filter settings demonstrate that layer architectures with larger filters in earlier convolutional layers and smaller filters in deeper convolutional layers optimize the accuracy result at high SNR. 
\begin{figure}[htb]
\centering
\includegraphics[width=0.8\columnwidth]{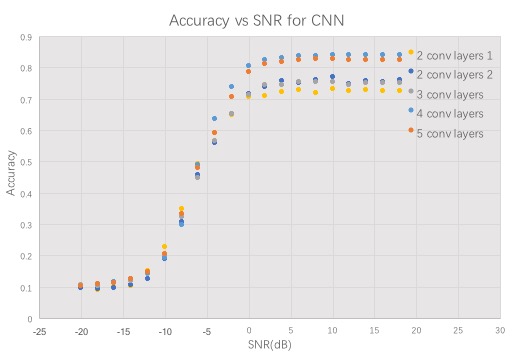}
\caption{Varying hyper-parameters in CNN. Accuracies at lower SNR are similar, the four-convolutional-layer architecture delivers an accuracy of 83.8\% at high SNR.}

\label{fig:4}
\end{figure} 

Next, we explore the optimal depth of CNN by increasing the number of convolutional layers from 2 to 5. We find that the best accuracy at high SNR is approximately 83.8\%. The best accuracy is obtained when using the four-convolutional-layer architecture as shown in Figure~\ref{fig:1}. This is a significant improvement of 8.8\% over the two-convolutional-layer model. Due to the fact that lower loss corresponds to higher accuracy, a smoothly decreasing loss indicates that the network is learning well as it does for the four-convolutional-layer model. When the neural network gets deeper, it becomes less likely for the validation loss to converge. For the five and six-convolutional-layer models, large loss vibrations appear early during training, which means that the minimum losses achieved by these neural networks are larger than that of the four-convolutional-layer model, which leads to the poor classification performance.

\subsection{Residual Network}
We find that combining a residual network with the original CNN architecture demonstrates similar performance as the pure CNN architecture. Similar to the result of CNN, the best performance of 83.5\% is achieved when we combine ResNet with a four convolutional layer neural network as shown in Figure~\ref{fig:2}. Recognition accuracy also starts to decrease when we combine ResNet with a network architecture that has more than four convolutional layers.
\subsection{Densely Connected Network}
\begin{figure}[htb]
\centering
\includegraphics[width=0.8\columnwidth]{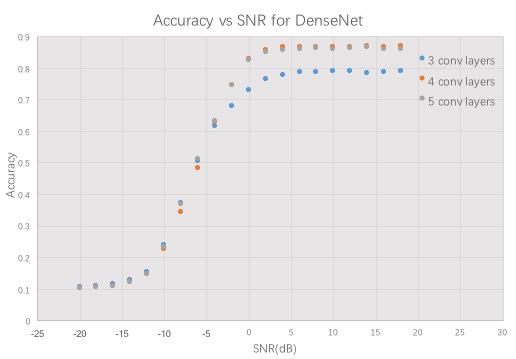}
\caption{Best Performance at high SNR is achieved with a four-convolutional-layer DenseNet.}
\label{fig:6}
\end{figure}

Because more densely connected blocks require a deeper neural network, which in our experiments did result in accuracy degradation, we implement DenseNet on CNN architectures with only one densely connected block. We start with a three convolutional layer DenseNet and keep adding convolutional layers into the network until the accuracy result starts to descend. We achieve a best accuracy of 86.6\% (see Figure~\ref{fig:6}) at high SNR using the four convolutional layer architecture shown in Figure~\ref{fig:3}.

\subsection{CLDNN}
\begin{figure}[htb]
\centering
\includegraphics[width=0.8\columnwidth]{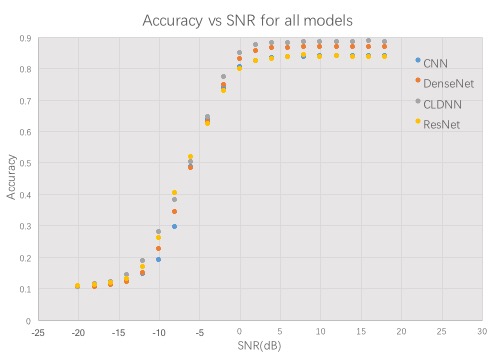}
\caption{Classification performance comparison between candidate architectures. CLDNN and DenseNet outperform other models with best accuracies of 88.5\% and 86.6\%, respectively. }
\label{fig:7}
\end{figure}

CLDNN has been widely used in recognition tasks that involve time domain signals like videos, speech, and images, as the inherent memory property leads to recognizing temporal correlations in the input signal. Recent work has also suggested the use of CLDNN for modulation recognition tasks \cite{deep-arch}. However, neither the network architecture nor the obtained accuracy results were clearly specified in~\cite{deep-arch}, and hence it was not feasible to reproduce these results and compare ours with. We applied the CLDNN architecture and compared the performance of CLDNN with results demonstrated by ResNet and DenseNet. We added an LSTM unit into the network after the convolutional part. We believe that the cyclic connections extract more relevant temporal features in the signal. The results of CLDNN - shown in Figure~\ref{fig:7} - do outperform other models. The accuracy at high SNR reaches 88.5\% and it is the highest among all tested neural network architectures. 
\begin{figure}[htb]
\centering
\includegraphics[width=0.8\columnwidth]{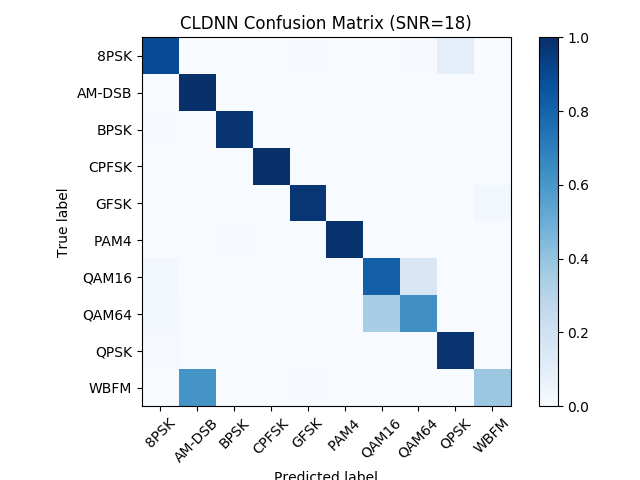}
\caption{The confusion matrix of CLDNN at SNR=18dB.}
\label{fig:matrix}
\end{figure} 

\begin{table}[htbp]
 \centering
 \begin{tabular}{lclcl}
  \toprule
  Misclassification  & Percentage(\%)\\
  \midrule
 8PSK/QPSK & 5.5 \\
 QAM16/QAM64 & 58.48 \\
 QAM64/QAM16 & 20.14 \\
 WBFM/AM-DSB & 59.6 \\
 WBFM/GFSK & 3.3 \\
  \bottomrule
 \end{tabular}
 \caption{Significant modulation type misclassification at high SNR for the proposed CLDNN architecture}
 \label{tbl:con}
\end{table}
In Figure~\ref{fig:matrix}, we show the classification results of the highest SNR case in a confusion matrix. There are two main discrepancies besides the clean diagonal in the matrix, which are WBFM being misclassified as AM-DSB and QAM16 being misclassified as QAM64. Details of the misclassification effects on accuracies are listed in Table~\ref{tbl:con}, where the number in the percentage column represents the percentage of the left hand side modulation type that is misclassified as the modulation type on the right hand side. A small portion of 8PSK samples are misclassified as QPSK and a small portion of WBFM samples are misclassified as GFSK; we expect that further optimizing the neural network architecture and possibly increasing the depth would lead to capturing these subtle feature differences. We further notice that QAM16 and QAM64 are likely to be misclassified as each other, since their similarities in the constellation diagram make the differentiation vulnerable to small noise in the signal. We expect that appropriate pre-processing of the input signal can help alleviate these large misclassification percentages. Large discrepancy also exists in WBFM classification which is likely to be recognized as AM-DSB. We believe that this discrepancy is probably due to the silence period where only carrier tone exists in the analog voice signal. 

\section{Discussion}\label{sec:comp}
By creating shortcuts between different layers, the ResNet and DenseNet architectures alleviate the vanishing gradient problem and promote feature reuse. By comparing the performances of ResNet and DenseNet in Figure~\ref{fig:7}, we notice that DenseNet demonstrates significantly better performance than ResNet by including more shortcut connections in the network, and therefore further strengthens feature propagation throughout the network. 
\begin{figure}[htb]
\centering
\includegraphics[width=0.8\columnwidth]{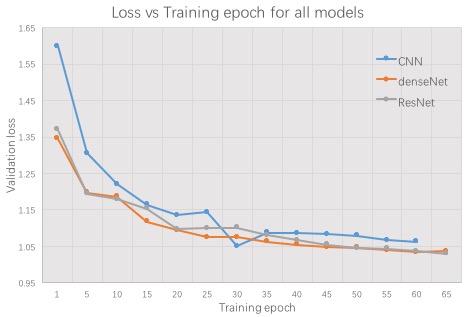}
\caption{Validation loss descents quickly in all three models,
but losses of DenseNet and ResNet reach plateau earlier than
that of CNN.}
\label{fig:5}
\end{figure} 

Although the ResNet and DenseNet architectures also suffer from accuracy degradation when the network grows deeper than the optimal depth, our experiments still show that when using the same network depth, DenseNet and ResNet have much lower convergence rates than plain CNN architectures. Figure~\ref{fig:5} shows validation errors of ResNet, DenseNet, and CNN of the same network depth with respect to the number of training epochs used. We can see that the ResNet and the DenseNet start at significantly lower validation errors and remain having a lower validation error throughout the whole training process, meaning that combining ResNet and DenseNet into a plain CNN architecture does make neural networks more efficient to train for the considered modulation classifcation task.

We finally applied the CLDNN architecture and obtained through it the best performance among all tested network architectures. We believe that the good performance of CLDNN is due to its long-term memory ability, which is suitable for the causality characteristic of time domain radio signals.

\section{Conclusion}
Multiple state of the art deep neural networks were applied for the radio modulation recognition task. We explored signal feature extraction by adding convolutional layers, various kinds of residual layers and recurrent layers to a deep neural network architecture. A Convolutional Long Short-term Deep Neural Network (CLDNN) was found to deliver the best classification architecture, which improves the accuracy by approximately 13.5\% upon the original CNN model introduced in~\cite{conv}. We believe that the causality of radio time domain signals leads to this improvement since a recurrent network is known to perform well for continuous acoustic signal processing tasks. The residual and densely connected networks (ResNet and DenseNet) also perform well although the best accuracy is limited by the depth of network, but they suggest that changing connections between layers - and specially creating shortcuts between non-consecutive layers - may produce better classification accuracy.

\bibliographystyle{IEEEtran}
\bibliography{refs.bib}

\begin{thebibliography}{1}
\providecommand{\url}[1]{#1}
\csname url@samestyle\endcsname
\providecommand{\newblock}{\relax}
\providecommand{\bibinfo}[2]{#2}
\providecommand{\BIBentrySTDinterwordspacing}{\spaceskip=0pt\relax}
\providecommand{\BIBentryALTinterwordstretchfactor}{4}
\providecommand{\BIBentryALTinterwordspacing}{\spaceskip=\fontdimen2\font plus
\BIBentryALTinterwordstretchfactor\fontdimen3\font minus
  \fontdimen4\font\relax}
\providecommand{\BIBforeignlanguage}[2]{{%
\expandafter\ifx\csname l@#1\endcsname\relax
\typeout{** WARNING: IEEEtran.bst: No hyphenation pattern has been}%
\typeout{** loaded for the language `#1'. Using the pattern for}%
\typeout{** the default language instead.}%
\else
\language=\csname l@#1\endcsname
\fi
#2}}
\providecommand{\BIBdecl}{\relax}
\BIBdecl

\bibitem{conv}
T.~J. O'Shea and J.~Corgan, ``Convolutional radio modulation recognition
  networks,'' \emph{CoRR}, vol. abs/1602.04105, 2016.

\bibitem{resnet}
K.~He, X.~Zhang, S.~Ren, and J.~Sun, ``Deep residual learning for image
  recognition,'' \emph{CoRR}, vol. abs/1512.03385, 2015.

\bibitem{densenet}
G.~Huang, Z.~Liu, and K.~Q. Weinberger, ``Densely connected convolutional
  networks,'' \emph{CoRR}, vol. abs/1608.06993, 2016.

\bibitem{CLDNN}
T.~N. Sainath, O.~Vinyals, A.~W. Senior, and H.~Sak, ``Convolutional, long
  short-term memory, fully connected deep neural networks,'' \emph{2015 IEEE
  International Conference on Acoustics, Speech and Signal Processing
  (ICASSP)}, pp. 4580--4584, 2015.

\bibitem{datagen}
T.~O'Shea and N.~West, ``Radio machine learning dataset generation with gnu
  radio,'' \emph{GNU Radio Conference}, vol.~1, no.~1, 2016.

\bibitem{adam}
D.~P. Kingma and J.~Ba, ``Adam: {A} method for stochastic optimization,''
  \emph{CoRR}, vol. abs/1412.6980, 2014.

\bibitem{deep-arch}
N.~E. West and T.~J. O'Shea, ``Deep architectures for modulation recognition,''
  \emph{CoRR}, vol. abs/1703.09197, 2017.

\end{thebibliography}










\end{document}